\documentclass[runningheads,a4paper]{llncs}

\usepackage{amssymb}
\usepackage{graphicx}
\usepackage{multirow}
\usepackage{placeins}

\usepackage{url}

\begin{document}

\mainmatter

\title{Multi-label Ferns for Efficient Recognition of Musical Instruments in Recordings}

\titlerunning{Multi-label Ferns for Efficient Recognition of Instruments}

\author{Miron B. Kursa\inst{1}
\and Alicja A. Wieczorkowska\inst{2}}
\authorrunning{Miron B. Kursa \and Alicja A. Wieczorkowska}

\institute{Interdisciplinary Centre for Mathematical and
Computational Modelling (ICM), University of Warsaw, Pawi\'{n}skiego 5A,
02-106 Warsaw, Poland\ \and
  Polish-Japanese Institute of Information Technology,
Koszykowa 86, \\ 02-008 Warsaw, Poland \\
\email{M.Kursa@icm.edu.pl},\
\email{alicja@poljap.edu.pl}\
}

\toctitle{Multi-label Ferns for Efficient Recognition of Musical Instruments in Recordings}
\tocauthor{Miron B. Kursa,
Alicja A. Wieczorkowska}
\maketitle

\begin{abstract}

In this paper we introduce multi-label ferns, and apply this technique for automatic classification of musical instruments in audio recordings.
We compare the performance of our proposed method to a set of binary random ferns, using jazz recordings as input data.
Our main result is obtaining much faster classification and higher F-score.
We also achieve substantial reduction of the model size.
\end{abstract}

\section{Introduction}

Music Information Retrieval (MIR) is a hot research topic last years \cite{MIR}, \cite{shen}, with quite a successful solving of such problems as automatic song identification through query-by-example, also using mobile devices \cite{shazam}, \cite{trackid}, and finding music works through query-by-humming \cite{midomi}.
Still, one of the unattainable goals of MIR research is automatic score extraction from audio recordings, which is especially difficult for polyphonic data \cite{her1}, \cite{kitahara}.
Multi-pitch tracking combined with assignment of the extracted notes to particular voices (instruments) is a way to approach score extraction.
Therefore, identification of instruments can be used to assign each note in a polyphonic and polytimbral sound to the appropriate instrument.
However, the recognition of all playing instruments from recordings in polyphonic environment is still a challenging and unsolved task, related to multi-label classification of audio data representing a mixture of sounds.

In our work, the target is to recognize all instruments playing in the analysed audio segment.
No initial segmentation nor providing external pitch is required.
The instruments identification is performed on short sound frames, without multi-pitch tracking.
In our previous works, we were using sets (which we called batteries) of binary classifiers to solve the multi-label problem \cite{ISMIS11}, \cite{macau} of identification of instruments in polyphonic environment.
Random forests \cite{Breiman} and ferns \cite{Ozuysal2007}, \cite{Ozuysal2008} were applied as classification tools.
Recently, we have shown that random ferns are a good replacement for random forests in music annotation tasks, as this technique offers similar accuracy while being much more computationally efficient \cite{BMC}.
In this paper we propose a generalized version of random ferns, which can natively perform multi-label classification.
Using real musical recording data, we will show that our approach outperforms a battery of binary random ferns classifiers in every respect: in terms of accuracy, model size and prediction speed.

\subsection{Background}

The difficulty level of automatic instrument recognition in audio data depends on the polyphony level, and on the preprocessing performed.
For single isolated sounds the instrument identification can even reach 100\% for a few classes, but it decreases to about 40\% for 30 or more classes \cite{her1}).
For polyphonic input even labelling of ground truth data is difficult, so mixes and single sounds are commonly applied to facilitate the research on polyphonic audio data.
The identification of instruments in polyphony is often supported with external provision of pitch data, but automatic multi-pitch tracking problem is addressed too \cite{Heittola}.
Another simplified approach aims at the identification of a predominant instrument \cite{Bosch}.
Multi-target identification of multiple instruments is performed as well, although this research is done on various sets of data, so the results cannot be directly compared.
This section presents a general view of the state of the art in this area, shows classification methods applied, and sketches the broad range of approaches used in this research.

The simplest polyphonic research case is instrument identification in duets (2 instruments) \cite{eggink}, \cite{Mike}, \cite{Vincent}, and the most complex one for symphonies, with high polyphony level (i.e. high number of instrument sounds played together).
In all cases the sound waves of instruments overlap, and so harmonic spectral components (partials) do, to a certain --- sometimes large --- extent.
One of the approaches consists in omitting overlapped partials
\cite{eggink}, resulting in about 60\% accuracy using Gaussian Mixture Models for duets from 5-instrument set.
Another interesting approach to multiple-instrument recognition is presented in \cite{Cont}; their approach was inspired by non-negative matrix factorization, with an explicit sparsity control.

Audio data are usually parametrized before further processing in the classification procedure, and pure data representing amplitude changes of a complex audio wave are rarely used.
Preprocessing consists in calculation of parameters describing audio signal, or (more often) spectral features.
Still, direct spectrum/template matching can be also for instrument identification, without feature extraction \cite{Mike}, \cite{Kashino}.
This approach can result in good accuracy; in \cite{Kashino}, 88\% was obtained for the polyphony of 3 instruments: flute, violin and piano, supported with integrating musical context into the system.

The accuracy of instrument identification usually drops with increasing polyphony level and number of instruments considered in the recognition procedure.
In \cite{kitahara}, 84.1\% was obtained for duets, 77.6\% for trios, and 72.3\% for quartets, using LDA (Linear Discriminant Analysis) based approach.
In \cite{Barbedo}, LDA yielded 60\% average precision for instrument pairs (300 pairs, 25 instruments), and much a higher recall of 86--100\%.
Other techniques used in multiple instrument identification include SVM (Support Vector Machine), decision trees, and k-nearest neighbour classifiers \cite{Essid}, \cite{Pardo}.
For the polyphony of up to four jazz instruments, the average accuracy of 53\% was obtained in \cite{Essid}, whereas \cite{Martins} obtained 46\% recall and 56\% precision for the polyphony of up to 4 notes for 6 instruments, based on spectral clustering, and PCA (Principal Component Analysis).
The research in \cite{Martins} aimed at sound separation, which is also performed as an intermediate step in automatic music transcription, and then each separated sound can be independently labelled.
Semi-automatic music transcription is addressed in \cite{kirch} through shift-variant non-negative matrix deconvolution (svNMD) and k-means clustering; the accuracy dropped below 40\% for 5 instruments, analysed in form of mixes.

\section{Data}

The data we use originate from various recordings, all recorded at 44.1kHz/16-bit, or converted to this format.
Testing is also performed on recordings, not on mixes.
This was possible because we used recordings especially prepared for research purposes, the original tracks for each instruments were available, and thus ground truth labeling was facilitated.
Both training and testing data were used as mono input, although some of them were originally recorded in mono or stereo format.
In the case of stereo data, mixes of the left and right channel (i.e. the average value of samples in both channels) were taken.

Sound parametrization was performed as a preprocessing in our research, for 40-ms frames.
Spectrum was calculated first, using Fourier transform with Hamming window, and various spectral features were extracted.
No pitch tracking was performed nor required as preprocessing.
Both training and testing data were labelled with the instrument or instruments playing in a given segment.
In the testing phase, the identification of instruments is performed on frame by frame basis, for consequent 40-ms frames, with 10 ms hop size (75\% overlap).

\subsection{Feature Set}
The feature vector consists of parameters describing properties of a 40-ms audio frame, and differences of the same parameters but calculated between for a 30 ms sub-frame starting from the beginning of the frame and a 30 ms sub-frame with 10 ms offset.
The features we used are mainly MPEG-7 low-level audio descriptors, are often used in audio research \cite{MPEG-7}, and other features applied in instrument recognition research.
The following 91 parameters constitute our feature set \cite{ISMIS11}, \cite{macau}:

\begin{itemize}
\item \textit{Audio Spectrum Flatness}, \textit{flat}$_{1}, \ldots,\; $\textit{flat}$_{25}$ --- features
parameter describing the flatness property of the power spectrum
within a frequency bin for selected bins; we used 25 out of 32 frequency
bands; 
\item \textit{Audio Spectrum Centroid} --- the power weighted average of the
frequency bins in the power spectrum, with coefficients scaled to an octave scale anchored at 1 kHz \cite{MPEG-7};
\item \textit{Audio Spectrum Spread} --- 
RMS (root mean square) of the deviation of
the log frequency power spectrum 
wrt. \textit{Audio Spectrum Centroid}
\cite{MPEG-7};
\item \textit{Energy} --- energy of the spectrum, in log scale;
\item \textit{MFCC} --- 13 mel frequency cepstral coefficients.
The cepstrum was calculated as the logarithm of
the magnitude of the spectral coefficients, and then transformed to the mel scale, reflecting properties of the human perception of frequency. 
24 mel filters were applied, and the results were transformed 
to 
12 coefficients, and the logarithm of the energy was taken as 
$13^{th}$ coefficient (0-order coefficient of MFCC)
\cite{MFCC};
\item \textit{Zero Crossing Rate}, where zero-crossing is a point where the sign of the sound
wave in time domain changes; 
\item \textit{Roll Off} --- the frequency below which an experimentally
chosen percentage (85\%) of the accumulated magnitudes of
the spectrum is concentrated; parameter originating from
speech recognition, applied to distinguish between voiced and unvoiced speech;
\item \textit{NonMPEG7 - Audio Spectrum Centroid} --- a linear scale version of \textit{Audio Spectrum Centroid};
\item \textit{NonMPEG7 - Audio Spectrum Spread} --- a linear scale version of \textit{Audio Spectrum Spread};
\item changes (differences) of the above features
    for a 30 ms sub-frame of the given 40 ms frame (starting from the beginning of this frame) and the next 30 ms sub-frame (starting with 10 ms offset);
\item \textit{Flux} --- the sum of squared differences between the magnitudes of the DFT points calculated for the starting and ending 30 ms sub-frames within the main 40 ms frame; this feature works on spectrum directly, not on its parameters.
\end{itemize}

\subsection{Audio Data \label{sec:adata}}

In our experiments we focused on wind instruments, typically used in jazz music.
Training data for clarinet, trombone, and trumpet were taken from three repositories of single, isolated sounds of musical instruments: McGill University Master Samples (MUMS) \cite{Opo_Wap}, The
University of Iowa Musical Instrument Samples (IOWA) \cite{IOWA}, and RWC Musical Instrument Sound Database \cite{RWC}.
Since so sousaphone sounds were available in these sets, we additionally used sousaphone sounds recorded by R. Rudnicki\cite{RR}.
Training data were in mono format in RWC data and for sousaphone, and in stereo for the rest of the data.
Training was performed on single sounds and mixes.
Our classifiers were trained to work on larger instrument sets, so additionally sounds of 5 other instruments were used in the training.
These were instruments also typical for jazz recordings: double bass, piano, tuba, saxophone, and harmonica. RWC, IOWA and MUMS repositories were used to collect these sounds.
The testing data were taken from the following jazz band stereo recordings by R. Rudnicki \cite{ISMIS11}, \cite{RR}:
\begin{itemize}
\item \textit{Mandeville} by Paul Motian,
\item \textit{Washington Post March} by John Philip Sousa, arranged by Matthew Postle,
\item \textit{Stars and Stripes Forever} by John Philip Sousa, semi-arranged by Matthew Postle --- Movement no. 2 and Movement no. 3.
\end{itemize}

These recordings contain pieces played by clarinet, trombone, trumpet, and sousaphone, which are our target instruments.

\section{Classification}
In the previous works, we have been solving the multi-label problem of recognising instruments with the standard binary relevance approach.
Namely, we were building a battery of binary models, each capable of detecting the presence or absence of a single instrument; for prediction, we were applying all the models to the sample and combining their predictions.

Unfortunately, this approach is not computationally effective, ignores the information about instrument-instrument interactions and requires sub-sampling of the training data to make balanced training sets for each battery member.
Thus, we attempted to modify the random ferns classifier used in our methodology to natively support multi-label classification.

\subsection{Multi-label Random Ferns}
Random ferns classifier is an ensemble of $K$ ferns, simple base classifiers equivalent to a constrained decision tree.
Namely, the depth of a fern ($D$) is fixed and the splitting criteria on a given tree level are identical.
This way, a fern has $2^{D}$ leaves and directs object $x$ into a leaf number $F(x)=1+\sum_{i=1}^{D} 2^{i-1}\sigma_{i}(x)\in 1..2^{D}$, where $\sigma_{i}(x)$ is an indicator variable for a result of the $i$-th splitting criterion.
We are using the rFerns implementation of random ferns \cite{JSS} which generates splitting criteria entirely at random, i.e. randomly selects both a feature on which the split will be done and the threshold value.
Also, rFerns builds a bagging ensemble of ferns, i.e. each fern, say $k$-th, is not directly build not on a whole set of objects but on a \textit{bag} $B_k$, a multiset of training objects created by random sampling with replacement the same number of objects as in the original training set.

The leaves of ferns are populated with \textit{scores} $S_k(x,y)$, indicating the confidence of a fern $k$ that an object $x$ falling into a certain leaf $F_k(x)$ belongs to the class $y$.
The scores are generated based on a training dataset $X^{t}=\{x^{t}_{1},x^{t}_{2},\ldots\}$, and are defined as
\begin{equation}
 e^{S_k(x,y)}=\frac{1+|L_k(x)\cap Y_k(y)|}{C+|L_k(x)|}
 \cdot
 \frac{C+|B_k|}{1+|Y_k(y)|},
\end{equation}
where $L_k(x)=\{x^t\in B_k : F_k(x)=F_k(x^t)\}$ is a multiset of training objects from a bag in the same leaf as a given object and $Y_k=\{x^t\in B_k : y\in Y(x^t)\}$ is a multiset of training objects from a bag that belong to a class $y$.
$Y(x)$ denotes a set of true classes of an object $x$, and is assumed to always contain a single element in a many-classes case; $C$ is the number of all classes.
The prediction of the whole ensemble for an object $x$ is
\begin{equation}
Y^{p}(x)=\arg\max_{y}\sum_{k=1}^{K} S_{k}(x,y).
\end{equation}

Our proposed generalisation of random ferns for multi-label classification is based on the observation that while the fern structures are not optimised to a given problem, the same set of $F_k$ functions can serve all classes rather than being re-created for each one of them.
In the battery classification, we create virtual \textit{not-class} classes to get a baseline score value used to decide whether class of a certain score value should be reported as present or absent.
With multi-class random ferns, however, we can incorporate this idea as a normalisation of scores so that the sign of their value will become meaningful indicator of a class presence.
We call such normalised scores \textit{score quotients} $Q_k(x,y)$, and define them as
\begin{equation}
 e^{Q_k(x,y)}=\frac{1+|L_k(x)\cap Y_k(y)|}{1+|L_k(x) \setminus Y_k(y)|}
 \cdot
 \frac{1+|B_k \setminus Y_k(y)|}{1+|Y_k(y)|}.
\end{equation}
The prediction of the whole ensemble for an object $x$ naturally becomes
\begin{equation}
Y^{p}(x)=\{y:Q_k(x,y)>0\}.
\end{equation}

\section{Experiments}
When preparing training data, we start with single isolated sounds of each target instrument.
After removing starting and ending silence \cite{ISMIS11},  each file  representing the whole single sound is normalized so that the RMS value equals one.
Then, we create the training set of sounds by mixing random 40 ms frames extracted from the recordings of 1 to 4 randomly chosen instruments; the mixing is done with random weights and the result is normalized again to get the RMS value equal to one.
Finally, we convert the sound into a vector of features by applying previously described sound descriptors.
The multi-label decision for such an object is a set of instruments which sounds were used to create the mix.
We have repeated this procedure 100 000 times to prepare our training set.

This set is used directly to generate the model with the multi-label random ferns approach.
When creating the battery of random ferns, we are splitting this data into a set of binary problems.
Each is devoted to one instrument and contains 3000 positive examples where this instrument contributed to the mix and 3000 negative when it was absent.

In both cases, we used $K=1000$ ferns and scanned depths $D=5,7,10,11,12$.
As the random ferns is a stochastic algorithm, we have replicated training and testing procedure 10 times.

Both models are tested on a true jazz recordings described in Section~\ref{sec:adata} and their predictions assessed with respect to the annotation performed by an expert.
The accuracy was assessed via precision and recall scores; these measures were weighted by the RMS of a given frame, in order to diminish the impact of softer frames which cannot be reasonably identified as their loudness approaches the noise level.
Our true positive score $T_{p}$ for an instrument $i$ is a sum of RMS of frames which are both annotated and classified as $i$.
Precision is calculated by dividing $T_{p}$ by the sum of RMS of frames which are classified as $i$; respectively, recall is calculated by dividing $T_{p}$ by the sum of RMS of frames which are annotated as $i$.

As a general accuracy measure we have used F-score, defined as a harmonic mean of such generalised precision and recall.

\section{Results}
\begin{figure}[p]
\centering
\includegraphics[width=\textwidth]{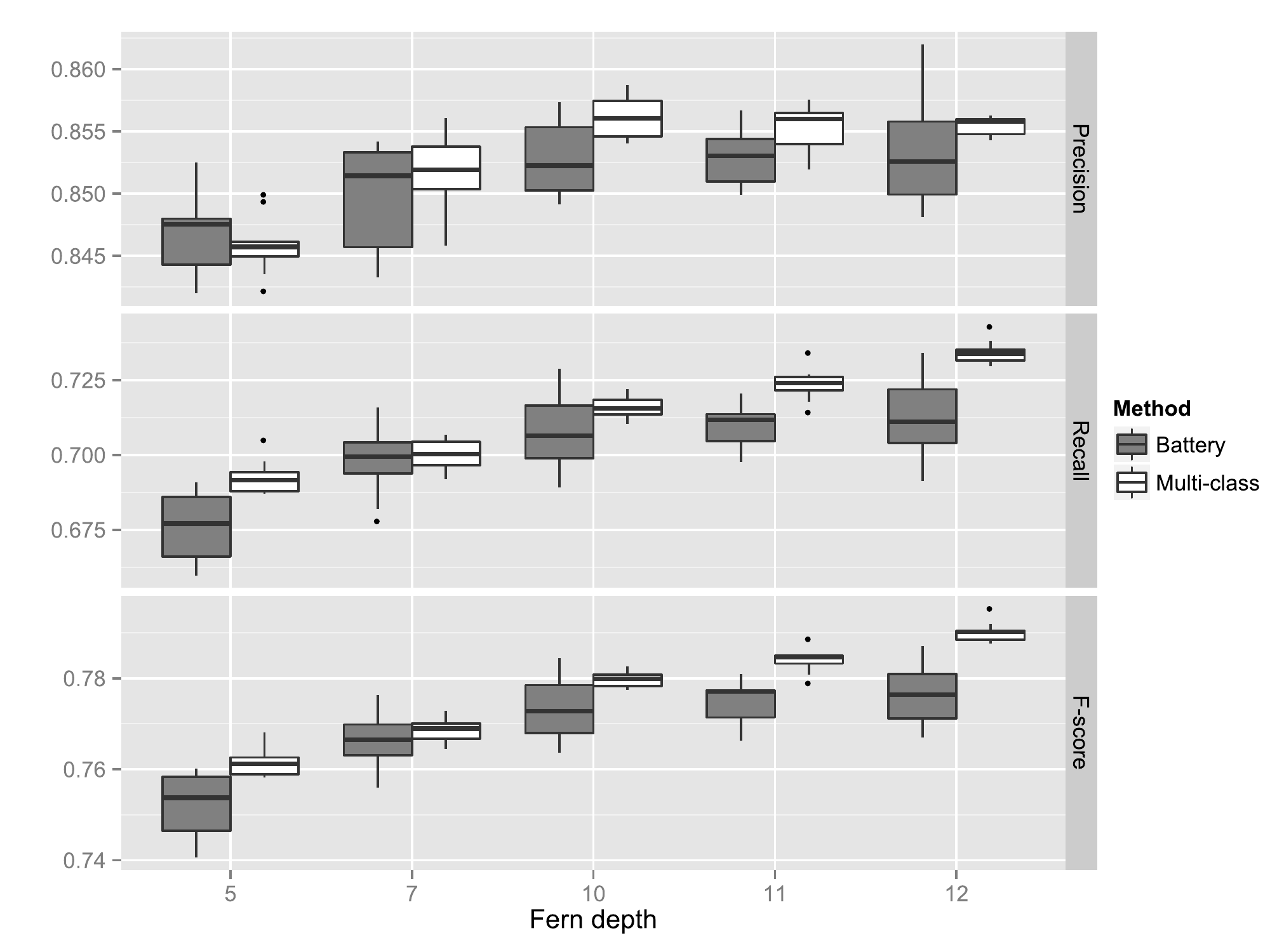}
\caption{Overall precision, recall and F-score for all the investigated jazz recordings and all the instruments for a battery of binary random ferns and for multi-label ferns.}
\label{fig:overall}
\end{figure}

The results of accuracy analysis are presented in a Figure~\ref{fig:overall}.
One can see that for fern depth larger than 7 the multi-label ferns archived both significantly better precision and recall that the battery classifier; obviously this also corresponds to a higher F-score.
The precision of both methods seems to stabilise for larger depth, while the recall and so F-score of multi-class ferns raise steadily and may be likely further improved.
The variation of the results is also substantially smaller for multi-class ferns, showing that the output of this approach is more stable and thus more predictable.

Table~\ref{tab:details} collects the sizes of created models and the speed with which they managed to predict the investigated jazz pieces.
One can see that the utilisation of multi-label ferns results in a substantially greater prediction speed, on average 7 times better than this achieved by the battery of binary ferns.

The difference between model sizes is less pronounced, with the battery actually producing a smaller model for fern depths 5 and 7 and only about 2 times larger for fern depth 12.

There is a negative correlation between the achieved F-score and both prediction speed and model size, though, with the fern depth controlling the speed-quality trade-off.
However, this way an user may use this parameter to flexibly adjust the model to the constraints of the intended implementation.

\begin{table}[p]
 \centering
\begin{tabular}{|c|c|c||c|c|}
 \hline
 & \multicolumn{2}{|c||}{Model size} & \multicolumn{2}{c|}{Prediction speed} \\
Fern depth & Battery & Multi-label & Battery & Multi-label \\
 \hline
5  & 6MB   & 13MB  & 54$\times$ & 359$\times$ \\
7  & 20MB  & 19MB  & 42$\times$ & 301$\times$ \\
10 & 143MB & 81MB  & 33$\times$ & 238$\times$ \\
11 & 284MB & 151MB & 30$\times$ & 216$\times$ \\
12 & 565MB & 292MB & 26$\times$ & 204$\times$ \\
 \hline
\end{tabular}
\vspace{2mm}
 \caption{Comparison of model size and prediction speed for a random ferns battery and multi-label random ferns. The speed is expressed as the total playing time of all investigated jazz recordings divided by the CPU time required to classify them. \label{tab:details}}
\end{table}

\section{Summary and Conclusions}

In this paper we introduce multi-label random ferns as a tool for automatic identification of musical instruments in polyphonic recordings of a jazz band.
The comparison of performance of multi-label random ferns and sets of binary ferns shows that the proposed multi-label ferns outperform the sets of binary ferns in every respect.
Multi-label ferns are much faster, achieve higher F-score, and the model size increase with increasing complexity also compares favorably with the set of binary random ferns.
Therefore, we conclude that multi-label random ferns can be recommended as a classification tools in many applications, not only for instrument identification, and this technique can also be applied on resource-sensitive devices, e.g. mobile devices.

\subsubsection*{Acknowledgments.}
This project was partially supported by the Research Center of PJIIT, supported by the Ministry of Science and Higher Education in Poland, and the Polish National Science Centre, grant 2011/01/N/ST6/07035.
Computations were performed at the ICM UW, grant G48-6.

\end{document}